\documentclass[onecolumn,aps,floatfix,superscriptaddress]{revtex4}
\usepackage[mathlines]{lineno}
\usepackage{graphicx}
\usepackage{amssymb}
\usepackage{amssymb}
\usepackage{pifont}
\usepackage{threeparttable}
\usepackage{natbib}

\usepackage{amsthm}
\usepackage{dsfont}
\usepackage{bbding}
\usepackage[utf8]{inputenc}

\usepackage{amsmath,amssymb,graphicx,bm,color,mathrsfs,verbatim,epstopdf,dcolumn,cancel}
\usepackage{bbold}
\usepackage[normalem]{ulem}
\useunder{\uline}{\ul}{}
\usepackage{booktabs}
\usepackage[noend]{algpseudocode}
\usepackage[ruled,vlined]{algorithm2e}
\usepackage{amssymb}
\usepackage{xcolor}
\usepackage{multirow}
\usepackage{array}
\usepackage[inline]{enumitem}
\usepackage{wrapfig}

\DeclareMathOperator{\SE}{SE}
\DeclareMathOperator{\SO}{SO}
\usepackage{caption}
\usepackage{float}
\usepackage{amsmath, amssymb}
\usepackage{array}
\makeatletter
\def\BState{\State\hskip-\ALG@thistlm}
\makeatother

\usepackage{multirow}
\usepackage[textsize=tiny]{todonotes}

\usepackage{hyperref}
\usepackage{cleveref}
\usepackage{pifont}

\begin{document}

\title{Clebsch-Gordan Transformer:\\ Fast and Global Equivariant Attention} 
\author{Owen Lewis Howell} \affiliation{Northeastern University: 360 Huntington Ave, Boston, MA 02115} \affiliation{The Boston Dynamics AI Institute: 145 Broadway, Cambridge, MA 02142}  \author{Linfeng Zhao} \affiliation{Northeastern University: 360 Huntington Ave, Boston, MA 02115} \author{Xupeng Zhu} \affiliation{Northeastern University: 360 Huntington Ave, Boston, MA 02115} \author{Yaoyao Qian} \affiliation{Northeastern University: 360 Huntington Ave, Boston, MA 02115} \author{Haojie Huang} \affiliation{Northeastern University: 360 Huntington Ave, Boston, MA 02115}\author{Lingfeng Sun} \affiliation{The Boston Dynamics AI Institute: 145 Broadway, Cambridge, MA 02142}  \author{Wil Thomason} \affiliation{The Boston Dynamics AI Institute: 145 Broadway, Cambridge, MA 02142} \author{Robert Platt} \affiliation{Northeastern University: 360 Huntington Ave, Boston, MA 02115} \affiliation{The Boston Dynamics AI Institute: 145 Broadway, Cambridge, MA 02142} \author{Robin Walters} \affiliation{Northeastern University: 360 Huntington Ave, Boston, MA 02115} \affiliation{The Boston Dynamics AI Institute: 145 Broadway, Cambridge, MA 02142}

\begin{abstract}
The global attention mechanism is one of the keys to the success of transformer architecture, but it incurs quadratic computational costs in relation to the number of tokens. On the other hand, equivariant models, which leverage the underlying geometric structures of problem instance, often achieve superior accuracy in physical, biochemical, computer vision, and robotic tasks, at the cost of additional compute requirements. As a result, existing equivariant transformers only support low-order equivariant features and local context windows, limiting their expressiveness and performance. This work proposes Clebsch-Gordan Transformer, achieving efficient global attention by a novel Clebsch-Gordon Convolution on $\SO(3)$ irreducible representations. Our method enables equivariant modeling of features at all orders while achieving ${O}(N \log N)$ input token complexity. Additionally, the proposed method scales well with high-order irreducible features, by exploiting the sparsity of the Clebsch-Gordon matrix. Lastly, we also incorporate optional token permutation equivariance through either weight sharing or data augmentation. We benchmark our method on a diverse set of benchmarks including n-body simulation, QM9, ModelNet point cloud classification and a robotic grasping dataset, showing clear gains over existing equivariant transformers in GPU memory size, speed, and accuracy.
\end{abstract}

\maketitle

\section{Introduction}
Transformer-based models have demonstrated effectiveness beyond language processing, showing strong performance in geometry-aware tasks such as robotics, structural biochemistry, and materials science \citep{wu2024pointtransformerv3simpler,goyal2023rvt,pan20213d,zeni2024mattergengenerativemodelinorganic,rhodes2025orbv3atomisticsimulationscale}. For instance, 3D robotic perception tasks ranging from segmentation to object matching process point clouds and LiDAR data using attention mechanisms. These tasks heavily rely on token-based representations, and their performance is often constrained by the number of tokens the model can effectively handle. 
AlphaFold~\citep{jumper2021highly}, for example, employs equivariant transformers to predict protein structures with unprecedented accuracy by explicitly leveraging $\SE{(3)}$ symmetries such as rotations and translations. 
However, implementing an equivariant neural network structure typically incurs significant computational overhead and increased inference time. As a result, most current approaches are limited to small symmetry groups or low-order representations~\citep{thomas2018tensorfieldnetworksrotation,fuchs2020se3transformers3drototranslationequivariant,moskalev2024se3hyenaoperatorscalableequivariant,liaoequiformer,satorras2022enequivariantgraphneural,kondor2018clebschgordannetsfullyfourier}.
Enabling fast, low-memory overhead equivariant operations over large context windows is essential to scaling robust and sample-efficient learning in geometry-aware domains.

Unfortunately, maintaining equivariance while modeling a global geometric context is challenging due to
the computational demands of processing high dimensional data at scale. There are essentially two components that contribute to the computational complexity of $E(3)$-equivariant transformers: the time and memory scaling of the transformer with the number of tokens, $N$, and the time and memory complexity on the maximum harmonic degree, $\ell$. 
Naively, a global equivariant attention mechanism will have $O(N^{2})$ token complexity and $O(\ell^{6})$ harmonic complexity~\citep{passaro2023reducingso3convolutionsso2}. By assuming only local attention, the token complexity can be reduced to $O(dN)$, where $d$ is the local context window, at the cost of discarding information about long range correlations. In addition, various approximation techniques have been used to reduce the harmonic complexity to $O(\ell^{3})$ \citep{luo2024enablingefficientequivariantoperations}. Recently, $\SE(3)$-Hyena achieved $O(N\log{N})$ computational complexity using long convolution~\citep{romero2021ckconv, poli2023hyenahierarchylargerconvolutional} in the Fourier domain. However, it only supports up to first-order irreducible representations of SO(3) (i.e., scalar and vectors), making it difficult to capture higher-degree angular dependencies and limiting its ability to represent more complex, structured geometric patterns. 
Moreover, SE(3)-Hyena is not permutation invariant, making it most suitable for point clouds with a natural ordering.

This raises the question: can we design a method with global equivariant attention and $\mathcal{O}(N \log N)$ token complexity, support for arbitrary orders of spherical harmonics with $\mathcal{O}( \ell^{3} )$ complexity, and permutation invariance?  This work addresses this challenge by introducing Clebsch-Gordon Convolution on $\SO(3)$ irreducible representations of arbitrary degree.  We also enforce or encourage permutation invariance through either weight sharing or data augmentation. Our contributions can be summarized as follows:
\begin{itemize}[leftmargin=*]
\item We generalize the SE(3)-Hyena method of~\citep{moskalev2024se3hyenaoperatorscalableequivariant} to include equivariant features of all types. By exploiting the sparsity of the Clebsch-Gordon matrix,our method achieves ${O}(L^{3})$ harmonic scaling. 
\item By applying our proposed attention in the graph spectral domain, we achieve permutation-equivariant global attention in ${O}(N \log N)$ time.
\item We benchmark our method on a diverse array of tasks, including robotics, computer vision, and molecular biochemistry. 
Our method outperforms current state of the art methods on all tasks, with considerable reduction in memory usage.
\end{itemize}

\begin{figure}[H]
\centering
\begin{minipage}{\textwidth}
\centering
\includegraphics[width=\textwidth]{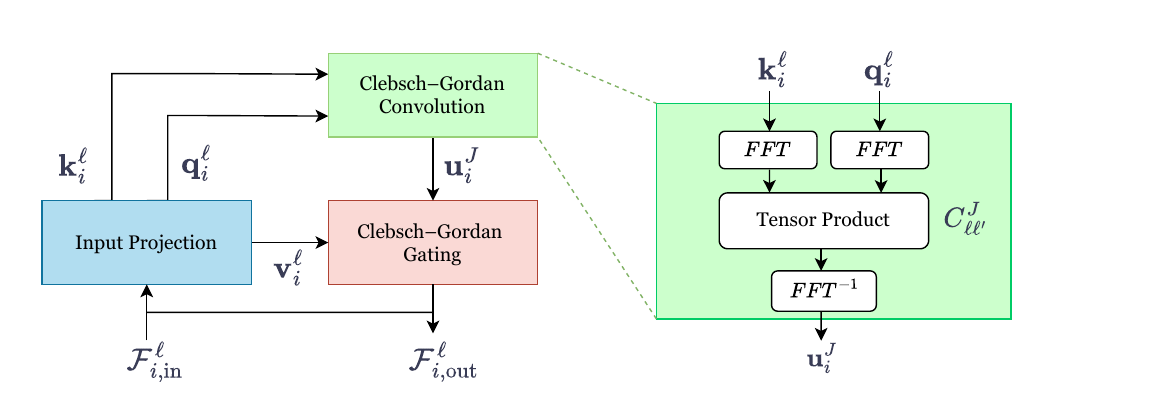}
\caption{
\small
Schematic of Clebsch-Gordon Convolution. Left: Inputs $f_{i}^{\ell}$ are projected into queries $q_{i}^{\ell}$, keys $k_{i}^{\ell}$, and values $v_{i}^{\ell}$ using an $SE(3)$-equivariant projection layer. Queries and keys are passed into a Clebsch-Gordon Convolution which outputs $u^{\ell}_{i}$ with which a tensor product of values is computed. Outputs are added with a residual connection. Right: Queries $q^{\ell}_{i}$ and keys $k^{\ell}_{i}$ are processed using Clebsch-Gordon Convolution. Queries and keys are first fast Fourier transformed, then a tensor product is applied. The output is fast Fourier transformed back.
}
\label{fig:graphic}
\end{minipage}
\end{figure}

\section{Related Work}



\paragraph{$\SE(3)$-Equivariance:} 
$\SE(3)$-equivariant neural networks have three main classes:
\begin{enumerate*}[label=(\arabic*)]
\item\label{rel.groupconv} those based on group convolution~\citep{cohen2016group}, which discretizes $\SE(3)$, transforms a convolutional filter according to each group element in the discretization, and lastly performs cross-correlation using the transformed convolutional filter \citep{escnn, epn, e2pn};
\item\label{rel.irrep} those based on irreducible (spherical Fourier) representations, which provide a compact representation of $\SO(3)$ signals at each point in the point cloud \citep{thomas2018tensorfieldnetworksrotation, brandstetter2022geometricphysicalquantitiesimprove, liaoequiformer, fuchs2020se3transformers3drototranslationequivariant, passaro2023reducing,liao2024equiformerv2improvedequivarianttransformer}, and
\item\label{rel.vector} those based on scalar, vector, and multivector representations \citep{deng2021vector,moskalev2024se3hyenaoperatorscalableequivariant,brehmer2023geometricalgebratransformer} 
\end{enumerate*}.
Compared with group convolution (class~\ref{rel.groupconv}), the irreducible spherical Fourier representation (class~\ref{rel.irrep}) is more compact and avoids discretization errors---among these works, Tensor Field Network (TFN)~\cite{thomas2018tensorfieldnetworksrotation} leverages the tensor product to propagate information between points, 
$\SE(3)$-Transformer~\cite{fuchs2020se3transformers3drototranslationequivariant} extends TFN by using attention in the Fourier domain for local information passing, and
ESCN~\cite{passaro2023reducing} proposes approximating the tensor product in $\SO(3)$ by that in $SO(2)$, significantly reducing computational complexity. Vector representation (class~\ref{rel.vector}) has limited expressiveness, while the irreducible representation improves expressiveness as its order increases~\cite{liaoequiformer}. This work achieves efficient $\SE(3)$-equivariance with high-order irreducible representations by introducing a linear-time attention mechanism based upon the vector long convolution introduced by~\cite{moskalev2024se3hyenaoperatorscalableequivariant}.


\begin{table}[t!]
\centering
\label{tab:method-comparison}
\footnotesize
\renewcommand{\arraystretch}{1.5} 
\setlength{\tabcolsep}{5pt}     
\begin{tabular}{|l|c|c|c|l|}
\hline
\textbf{Model} & \textbf{\shortstack{Global \\ Attn.}} & \textbf{\shortstack{Perm. \\ Equiv.}} & \textbf{\shortstack{Token \\ Complex.}} & \textbf{\shortstack{Harmonic \\ Complex.}} \\ \hline \hline
SEGNN           & \ding{55}	 & \checkmark & $\mathcal{O}(dN)$      & $\mathcal{O}(L^6)$ \\ \hline
SE(3)-Transformer & \ding{55}	 & \checkmark & $\mathcal{O}(dN)$      & $\mathcal{O}(L^6)$ \\ \hline
Equiformer-v2   & \ding{55}	 & \checkmark & $\mathcal{O}(dN)$      & $\mathcal{O}(L^3)$ \\ \hline
SE(3)-Hyena     & \checkmark & \ding{55}	 & $\mathcal{O}(N\log N)$ & Type 0 \& Type 1 only \\ \hline \hline
\textbf{Ours}   & \checkmark & \checkmark/learned & $\mathcal{O}(N\log N)$ & $\mathcal{O}(L^3)$ \\ \hline
Theoretical Ideal & \checkmark & \checkmark & $\mathcal{O}(N\log N)$ & $\mathcal{O}(L^2\log L)$ \\ \hline
\end{tabular}
\caption{Key properties of equivariant attention mechanisms. $N$: number of tokens, $d$: average graph degree, $L$: maximum harmonic degree.}
\end{table}

\paragraph{Subquadratic Attention:}
Several linear-time attention mechanisms have been proposed to overcome the quadratic time and memory complexity of standard Transformer architectures. Reformer~\cite{kitaev2020reformer} utilizes locality-sensitive hashing (LSH) to approximate self-attention.~\cite{choromanski2021rethinking} replaces the standard softmax attention with a kernel-based approximation called FAVOR+ (Fast Attention Via positive Orthogonal Random features), achieving linear time and space complexity. Nyströmformer~\cite{xiong2021nystromformer}, leverages the Nyström method to approximate the self-attention matrix using a set of landmark points. Linformer~\cite{guo2024logformer} addresses the quadratic memory and computation bottleneck of standard Transformers by approximating self-attention with low-rank projections. These methods enable LLM transformers to process much longer sequences than previously feasible.

\section{Background}


\textbf{Attention.}
Attention is a data-dependent linear map~\citep{vaswani2023attentionneed} describing the pairwise interaction between tokens in a transformer's input context.
Let $q_{i}$, $k_{i}$ and $v_{i}$ be linear projections of input. 
Attention is defined $\text{Attn}( q, k , v ) = \text{softmax}[ \alpha(q,k) ] v$ where $\alpha(q,k)_{ij} = q^{T}_{i}k_{j}$ is the attention matrix. The computation of $\alpha$ scales quadratically in the input size, which is the main bottleneck in the transformer architecture in large models. Numerous methods attempt to compute $\alpha(q,k)$ faster using some numerical approximation.


\textbf{Equivariant Attention and Message Passing.} Numerous equivariant methods attempt to generalize attention to process equivarient features. The two most common forms are equivariant attention, or equivariant message passing ~\citep{brandstetter2022geometricphysicalquantitiesimprove}. In the standard setup, the $i$-th graph node is located at position $x_{i} \in \mathbb{R}^{3}$ has features $f_{i}^{\ell}$ where the index $\ell$ specifies feature type. Attention and message passing then attempt to process information via update rules
\begin{align}
& \text{Attention: } f^{\ell}_{out, i} = W_{V}^{\ell\ell}f_{in, i}^{\ell} + \sum_{k=0}^{L} \sum_{ j \in \mathcal{N}_{i}  } \alpha_{ij} W^{\ell k}_{V}(x_{i} - x_{j} ) f^{k}_{in, j} \\
& \text{Message Passing: } f^{\ell}_{ out, i} = \phi( f^{\ell}_{in, i}, \sum_{j \in \mathcal{N}_{i}} m_{ij} ), \enspace m_{ij} = \psi( f^{\ell}_{in,i} , f^{\ell'}_{in,j} , || x_{i} - x_{j} || )
\end{align}
where $\mathcal{N}_{i}$ is some neighborhood of points around point $j$. Memory constraints force methods like \citep{fuchs2020se3transformers3drototranslationequivariant, passaro2023reducingso3convolutionsso2,satorras2022enequivariantgraphneural,brandstetter2022geometricphysicalquantitiesimprove,thomas2018tensorfieldnetworksrotation} to restrict to small neighborhoods $\mathcal{N}_{i}$ of size less than 50. We show in Sec. \ref{Section:Experiments} that decreasing the size of the local context window can lead to significant changes in performance.
A recent method~\citep{moskalev2024se3hyenaoperatorscalableequivariant} showed that for invariant and vector convolutions adding global context can improve model performance. We extend this work to equivariant features of all types.

\section{Method}

\newcommand{\tqueries}{\ensuremath{Q_{F^n}}}
\newcommand{\tkeys}{\ensuremath{K_{F^n}}}
\newcommand{\tvalues}{\ensuremath{V_{F^n}}}

Let $F^n=\lbrace f^{\ell}_{i} \rbrace_{i=1}^{n}$ be a set of $n$ input features to an $\SO(3)$-equivariant transformer, where $\SO(3)$ acts upon each feature $f^{\ell}_{i} \in \mathbb{R}^{(2\ell+1)\times m_{\ell}}$ via its $\ell^\text{th}$ irreducible representation.
We denote the multiplicity (i.e., channel dimension) of the input of type $\ell$ as $m_\ell$.
From the features in $F^n$, we encode queries, \tqueries, keys, \tkeys, and values, \tvalues, as:
\begin{align*}
q^{\ell}_{i} = W_{Q}^{\ell}( f^{\ell'}_{i} ), \quad  k^{\ell}_{i} = \sum_{\ell'} W_{K}^{\ell}( f^{\ell'}_{i} ), \quad v^{\ell}_{i} = W_{V}^{\ell}( f^{\ell'}_{i} )
\end{align*} 
where $W_{Q}^{\ell}$, $W_{K}^{\ell}$, and $W_{V}^{\ell}$ are learnable equivariant mappings converting type $\ell'$ features of multiplicity $m'_{\ell}$ into type $\ell$ features of multiplicity $m_{\ell}$. Our proposed method is agnostic to the particular equivariant encoders used; see~\cref{Appendix: Encoding Ablation} for more details. We seek to compute self-attention 
over \tqueries, \tkeys, and \tvalues{} in linear time, scaling to global context, and---unlike earlier work~\citep{moskalev2024se3hyenaoperatorscalableequivariant}--- remaining compatible with equivariant features of any type. Our proposed method extends the core idea of~\citet{poli2023hyenahierarchylargerconvolutional}, building upon the vector long convolution introduced by~\citet{moskalev2024se3hyenaoperatorscalableequivariant}.

\subsection{Clebsch-Gordon Convolution}

We structure our attention mechanism as follows:
first, inspired by~\citet{moskalev2024se3hyenaoperatorscalableequivariant}, we define the following operation, where $C^J_{\ell \ell'}$ is the Clebsch-Gordan matrix projecting from features of type $\ell \otimes \ell'$ onto features of type $J$:
\begin{align}\label{EquiConv}
( q^{\ell} \star k^{\ell'} )^{J}_{i} = C^{J}_{\ell \ell'} \sum_{j=1}^{N} q^{\ell}_{j} \otimes k^{\ell'}_{i-j}
\end{align}
which takes as input features of types $\ell$ and $\ell'$ and outputs features of type $J$. 
If $q^{\ell}_{i} \in \mathbb{R}^{ (2\ell +1 ) m_{\ell} }$ and $k^{\ell}_{i} \in \mathbb{R}^{ (2\ell +1 ) m_{\ell} }$ the resultant tensor product has dimension $( q^{\ell} \star k^{\ell'} )^{J}_{i} \in \mathbb{R}^{ (2J+1)m_{\ell}m_{\ell'}} $. 
In practice, we found that using multiple heads (which do not interact during the tensor product) led to better performance; see \cref{Appendix: Model Ablations} for further discussion. Operations of the form $C^{J}_{\ell \ell} q^{\ell} \otimes k^{\ell'}$ are ubiquitous in machine learning, making their fast computation a subject of great research interest. For the special case when the keys are spherical harmonic outputs, i.e., $k^{\ell} = Y^{\ell}( \hat{n} )$, \citet{passaro2023reducingso3convolutionsso2} used a group theoretic decomposition to reduce $\SO(3)$ operations into $\SO(2)$ operations. 
\citet{luo2024enablingefficientequivariantoperations} generalized this idea and used the Gaunt tensor product coefficients to reduce the tensor product computation to a highly tractable two dimensional Fourier transformation for general input features. 
We compute the tensor product in~\cref{EquiConv} using a slight modification of the methods proposed in \citep{luo2024enablingefficientequivariantoperations}; see  \cref{Appendix: Additional Experiments} for details. The operation in~\cref{EquiConv}, picks the type $J$ output out of the tensor product. 
By definition of the Clebsch-Gordon matrix $C^J_{\ell\ell'}$, the tensor product of type $\ell$ and type $\ell'$ features decomposes as
\begin{align}
( q^{\ell}_{i} \otimes k^{\ell'}_{i-j} ) = \bigoplus_{J} C^{J}_{\ell \ell'} (q^{\ell}_{i} \otimes k^{\ell'}_{i-j} )^{J}
\end{align}
where $\bigoplus$ is the direct sum of vector spaces. To allow all query and key types to interact, we want to compute, for each $J$, $ \hat{u}_{i}^{J} = \sum_{\ell \ell' } ( q^{\ell}_{i} \star k^{\ell'}_{i} )^{J} $. Following the notation of~\citet{luo2024enablingefficientequivariantoperations}, let $\tilde{q}_{i} = [q_{i}^{0},q_{i}^{1},...,q_{i}^{L}]$ be the stack of all query vectors containing irreducibles of up to degree $L$ and let $\tilde{k}_{i} = [k_{i}^{0},k_{i}^{1},...,k_{i}^{L}] $ be the stack of all keys containing irreducibles of up to degree $L$. 
The full tensor product of these features for output type $J$ is given by
\begin{align*}
u_{i}^{J} = ( \tilde{q}_{i} \star \tilde{k}_{i} )^{J} = \sum_{\ell=1}^{L} \sum_{\ell'=1}^{L} ( q_{i}^{\ell} \star k_{i}^{\ell'} )^{J} 
\end{align*}
Naively retaining all output-irreducible types of the Clebsch-Gordan tensor product up to type $L$ requires $O(L^{3})$ 3D matrix multiplications, for a total complexity of ${O}(L^{6})$.
We then define the full convolution as $\tilde{u}^{J}_{i} = ( \tilde{q}_{i} \star \tilde{k}_{i} )$. This convolution computation can be simplified in two ways.
The Fourier transform $\hat{u}^{\ell}_{q}$ of $u^{\ell}_{i}$ over the spatial index can be written as $\hat{u}^{J}_{i} = C^{J}_{\ell \ell'} \hat{q}^{\ell}_{i} \otimes \hat{k}^{\ell'}_{i}$ which is a matrix multiplication in Fourier space. Using the Fast Fourier Transform, the computation of $\hat{q}^{\ell}$ and $\hat{k}^{\ell}$ can be done in time $O( N \log N)$. 
We further consider both intra-channel and inter-channel tensor products; see~\cref{Appendix: Model Ablations} for additional ablation studies.

\subsubsection{Invariant Gating}
A key aspect of the transform proposed in~\citep{moskalev2024se3hyenaoperatorscalableequivariant} is its non-linear data dependent gating.
Accordingly, after obtaining $\hat{u}^{J}_{q}$, we compute a set of invariant features $I^{\ell} = \gamma^{\ell}( \hat{u}^{0},\hat{u}^{1}, \hat{u}^{2}, ...  )$ with one $I^{\ell}$ for each irreducible type $\ell$. We evaluated a variety of encoder types for $\gamma^{\ell}$; see~\cref{Append: Gating Abalation} for ablation studies. The gating $I^{J}$ is of dimension $N \times m_{J}$. We then apply softmax gating $u^{J}_{i}  \rightarrow \sigma( I^{\ell} ) u^{J}_{i} $ and combine the resultant gated features $u^{J}_{i}$ with the values via another tensor product and Clebsch-Gordon matrix projection
\begin{align}\label{Attention-Computation}
f_{i, out}^{J} = \sum_{\ell \ell} C^{J}_{ \ell \ell' } u^{\ell}_{i} \otimes v^{\ell'}_{i}
\end{align}
Note that the second multiplication,~\cref{Attention-Computation} is done in real space rather than Fourier space. 
The idea of switching between real and Fourier space when computing attention is a key idea developed by~\citet{poli2023hyenahierarchylargerconvolutional, stachenfeld2020graphnetworksspectralmessage}, allowing the model to fuse both global and local information. 
Lastly, we apply an equivariant MLP to reduce the multiplicity dimension back down to that of the input $ \hat{f}^{\ell}_{i} \rightarrow \text{MLP}( \hat{f}^{\ell}_{i}  ) $. We then add the attention features to the input features as a residual via
\begin{align*}
f^{\ell}_{i, out} = f^{\ell}_{i, in} + \text{MLP}( f^{\ell}_{i, in}  )  
\end{align*}
By subtracting off and re-adding the mean of inputs $f^{\ell}_{i}$, the outputs $\hat{f}^{\ell}_{i, out}$ are be fully $\SE(3)$-equivariant.


\begin{wrapfigure}{l}{0.52\textwidth}
\vspace{-10pt}
\begin{minipage}{\linewidth}
\footnotesize
\begin{algorithm}[H]
\DontPrintSemicolon
\SetKwInOut{KwIn}{\textbf{Input}}
\SetKwInOut{KwOut}{\textbf{Output}}
\SetAlgorithmName{Algorithm}{}{}
\caption{Clebsch-Gordon Convolution}
\KwIn{Input signal $f_{i,in}^{\ell}$}
\KwOut{Output Attention $f_{i,out}^{J}$}
\BlankLine
\nl \textbf{Encode:} $q^{\ell}_{i} = W_{Q}^{\ell}(f_{i}^{\ell'}), k^{\ell}_{i} = W_{K}^{\ell}(f_{i}^{\ell'}), v^{\ell}_{i} = W_{V}^{\ell}(f_{i}^{\ell'})$\;
\nl \textbf{FFT:} $q^{\ell}_{i}, k^{\ell}_{i} \rightarrow \hat{q}^{\ell}_{i}, \hat{k}^{\ell}_{i}$\;
\nl \textbf{Tensor Product:} $u^{J}_{i} = C^{J}_{\ell\ell'}\hat{q}_{i}^{\ell} \otimes \hat{k}_{i}^{\ell'}$ \;
\nl \textbf{Inverse FFT:} $\hat{u}^{\ell}_{i} \rightarrow u^{\ell}_{i}$\;
\nl \textbf{Tensor Product:} $f_{i,out}^{J} = C^{J}_{\ell\ell'}u_{i}^{\ell} \otimes v_{i}^{\ell'}$ \;
\nl \Return{$f_{i,out}^{J}$}\;
\end{algorithm}
\end{minipage}
\end{wrapfigure}

In practice, we found that using a head dimension of $4$ or $8$ with a maximum harmonic of $L=5$ of $L=6$ and a channel dimension of $8$ or $16$ was optimal. 
See~\cref{Appendix: Feature Cross Attention} for ablations on model parameters. In general, we found that using a fixed local context window allowed for much greater performance. Specifically, out full output features are
\begin{align*}
f^{\ell}_{i, out} = \mathcal{F}^{CG}( f^{\ell}_{i,in} ) + \mathcal{F}^{SGNN}( f^{\ell}_{i,in} )
\end{align*}
where $\mathcal{F}^{CG}( f^{\ell}_{i,in} )$ is the output of the Clebsch-Gordon convolution and $\mathcal{F}^{SGNN}( f^{\ell}_{i,in} )$ is a standard equiformer layer~\citep{passaro2023reducingso3convolutionsso2}, with a fixed local context window. We conjecture using both global and local attention that this allows the model to better process both local and global information. This idea is inspired by \citet{han2024demystifymambavisionlinear}, where it is conjectured that state space models combined with some small local attention can capture both global and local features.

\subsection{Architecture}

Figure \ref{fig:graphic} shows the full flow of our proposed attention block. Specific choices for architecture for experiments is discussed more in \ref{Appendix: Model Ablations}.

\subsection{Sparsity of the Clebsch-Gordon Matrix}\label{Sparsity}

\begin{wrapfigure}{l}{0.4\textwidth}
\centering
\includegraphics[width=\linewidth]{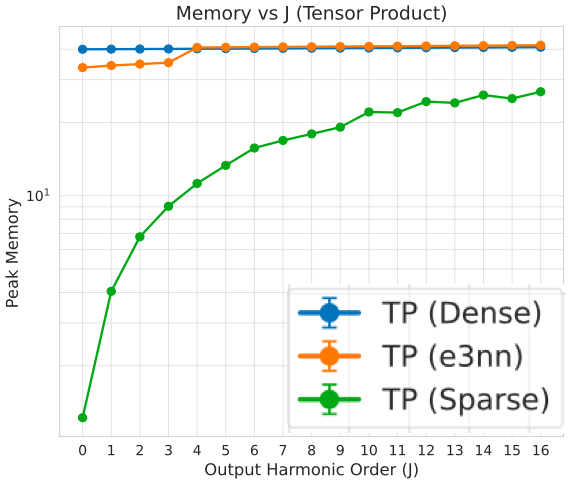}
\caption{
\small    
Memory usage versus the number of irreducible representations $J$ for our tensor product attention mechanism. We compare with dense matrix multiplication and e3nn\cite{geiger2022e3nneuclideanneuralnetworks}.  }
\label{fig:memory_vs_J}
\end{wrapfigure}

We now turn to improving the performance of our proposed attention mechanism by exploiting sparsity properties of the Clebsch-Gordon matrix. Consider the operation defined by$u^{J} = \sum_{\ell \ell'} C^{J}_{\ell \ell'} q^{\ell} \otimes k^{\ell'}$. The naive cost of this operation is $\mathcal{O}( (2J+1)(2\ell+1)(2\ell'+1) )$.
Thus, the total computation of $u^{\ell}$ for each $\ell$ is naively ${O}( J \ell \ell' ) = J L^{4}$ where $L$ is the maximum harmonic used. Fortunately, this neglects the sparsity of the matrix $C^{ J }_{\ell \ell'}$.
Specifically, $C^{J}_{\ell \ell'}$ is a tensor of size $(2J+1) \times (2\ell +1 ) \times (2\ell'+1)$ with most elements equal to zero.
To see this, let $|\ell m \rangle$ be the basis for the $\ell$ representation. 
Then, the Clebsch-Gordon coefficients $C^{JM}_{\ell m \ell' m'}$ satisfy
\begin{align*}
|JM \rangle = \sum_{mm'} C^{JM}_{\ell m \ell' m'} |jm \rangle \otimes |j'm' \rangle
\end{align*}
because $J^{2}$ and $J_{z}$ can be simultaneously diagonalized, it is always possible to find a basis (e.g., the $z$-basis is standard convention in physics) where $J_{z} | j m \rangle = m |j m \rangle $ applying this relation to the definition of the Clebsch-Gordon coefficients, we have that
\begin{align*}
J_{z} |JM \rangle  = M |JM \rangle \implies M |JM \rangle = \sum_{mm'} C^{JM}_{\ell m \ell' m'} (m + m ') |\ell m \rangle \otimes |\ell'm' \rangle
\end{align*}
Ergo, the matrix elements $C^{JM}_{\ell m \ell' m'}$ are non-zero only when $ M = m + m'$.
In physicists language, the sparsity of $C^{JM}_{\ell m \ell' m'}$ is a \emph{selection rule} that follows from conservation of the z-component of angular momentum.
Thus, the total number of non-zero elements in $C^{J}_{\ell\ell'}$ is $(2\ell +1 )(2\ell' + 1)$---much smaller than the naive estimate of $(2 J+1) \times (2\ell +1 ) \times (2\ell' + 1) $. 
Exploiting this fact, we can compute the tensor product in time dependent only on the input sizes. Parity symmetry further requires that any $C^{JM}_{\ell m, \ell' m'}$ where $\ell + \ell' + J$ is odd is identically zero; it also constrains some elements to be redundant. We compare the sparse method with the Gaunt tensor product method of~\citet{luo2024enablingefficientequivariantoperations} and the e3nn implementation~\citep{geiger2022e3nneuclideanneuralnetworks} in \cref{Appendix: Model Ablations}. 

\subsection{Special Case: Vector Long Convolution}

We show that our method recovers the vector long convolution method proposed in~\citep{moskalev2024se3hyenaoperatorscalableequivariant} as a special case. 
This vector long convolution is a fast $O(N \log N)$ attention mechanism for queries, keys and values of type 1. Let $q_{i}$ and $k_{i}$ be type 1 queries and keys. The vector convolution from~\citep{moskalev2024se3hyenaoperatorscalableequivariant} is defined as
\begin{align}\label{Vector Convolution}
( q \star k )_{i} = \sum_{j} q_{j} \times k_{i-j}
\end{align}
where $\times$ denotes the standard vector cross product. 
Elementwise, the vector convolution has $( q \star k)_{i a} = \epsilon_{abc} \sum_{j} q_{jb} k_{(i-j) c }$. The vector convolution $q \star k$ transforms in the vector representation of $SO(3)$. 
\Cref{Vector Convolution} is equal to the expression $( q \star k)_{i} = C^{1}_{11} \sum_{j} q_{j} \otimes k_{(i-j)}$. To see, this note that the tensor $C^{1}_{11}$, which is of dimension $(3 \times 3 \times 3)$ can be written out as a $(3 \times 9)$ matrix with elements given by
\begin{align*}
\footnotesize
C^1_{11} =
\begin{bmatrix}
0 & 0 & 0 & 0 & 0 & \frac{1}{\sqrt{2}} & 0 & \frac{1}{\sqrt{2}} & 0 \\
0 & 0 & \frac{1}{\sqrt{2}} & 0 & -\frac{1}{\sqrt{2}} & 0 & \frac{1}{\sqrt{2}} & 0 & 0 \\
0 & \frac{1}{\sqrt{2}} & 0 & \frac{1}{\sqrt{2}} & 0 & 0 & 0 & 0 & 0
\end{bmatrix}
\end{align*}
Note the for two 3-vectors $q$ and $k$, this can be written as $C^{1}_{11}( q \otimes k ) = \frac{1}{\sqrt{2}} ( q \times k)$. Thus, for any three vectors $C^{1}_{11}( q \otimes k ) = \frac{1}{\sqrt{2}} ( q \times k) $ is proportional to the standard cross product. Thus, we have that
\begin{align*}
( q \star k)_{i a} = \epsilon_{abc} \sum_{j} q_{jb} k_{(i-j) c } = \sqrt{2} C^{1}_{11} \sum_{j} q_{j} \otimes k_{(i-j)}
\end{align*}
which reduces to our method for input types $(1,1)$ and output type $1$. Thus, the method proposed in \citep{moskalev2024se3hyenaoperatorscalableequivariant} can be seen as a special case of our method when tensor product input features are restricted to be pairs of invariant or vector features.




\section{Experiments}\label{Section:Experiments}

\subsection{Baselines}

We compare our method with state of the art baselines for point cloud processing. The SE(3)-transformer \citep{fuchs2020se3transformers3drototranslationequivariant} is an equivariant attention mechanism based on Tensor Field Networks \citep{thomas2018tensorfieldnetworksrotation}. Equiformer v2 \citep{liao2024equiformerv2improvedequivarianttransformer} uses a convolutional trick from \citet{passaro2023reducingso3convolutionsso2} to reduce the harmonic complexity to $O(L^{3})$ from $O(L^6)$. Fused SE(3)-transformer implements the method of \citet{fuchs2020se3transformers3drototranslationequivariant} using fused kernels for decreased computational overhead. SE(3)-Hyena \citep{moskalev2024se3hyenaoperatorscalableequivariant} uses a modification of the Hyena architecture to do global linear time attention on invariant (type $0$) and vector (type $1$) features. The use of only invariant and vector features significantly limits model expressivity.

We benchmark our method on the ModelNet40 \citep{wu20153d} classification task, Nbody trajectory simulation \citep{fuchs2020se3transformers3drototranslationequivariant}, QM9 \citep{ramakrishnan2014quantum}, and a custom robotic grasping dataset. See \cref{Appendix: Additional Experiments} for all experiments.

\subsection{Nbody Simulation}


\begin{table}[H]
\centering
\label{tab:nbody-results}
\resizebox{\textwidth}{!}{%
\begin{tabular}{|l|c|c|c|c|c|c|c|c|c|}
\hline
\textbf{Model} & \textbf{Linear} & \textbf{Ours} & \textbf{Ours w/o local} & \textbf{SE(3)-Hyena} & \textbf{Set Trans.} & \textbf{SE(3)-Trans.} & \textbf{SEGNN} & \textbf{EGNN} & \textbf{TFN} \\
\hline
MSE $x$ & 0.0691 & \textbf{0.0041}$\pm$0.0003 & 0.0050$\pm$0.0003 & 0.0071 (0.0018) & 0.0139 & 0.0076 (0.0076) & 0.0048 (0.0056) & 0.0070 & 0.0150 \\
\hline
$\Delta$EQ & --- & 0.0000096 & 0.000010 & 0.00011 & 0.167 & 0.00000032 & NR & NR & NR \\
\hline
MSE $v$ & 0.261 & \textbf{0.0065}$\pm$0.0002 & 0.0075$\pm$0.0003 & 0.0071$\pm$0.0007 & 0.101 & 0.075 (0.075) & NR & NR & NR \\
\hline
$\Delta$EQ & --- & 0.00000048 & 0.00000052 & 0.0000012 & 0.370 & 0.00000063 & NR & NR & NR \\
\hline
\end{tabular}%
}
\caption{N-body simulation results ($N=5$ particles). Mean squared error for position ($x$) and velocity ($v$) prediction. Literature values in parentheses. NR = not reported.}
\end{table}

We first test on method on the Nbody simulation task, described in \citet{fuchs2020se3transformers3drototranslationequivariant}. In the simulation in \citet{fuchs2020se3transformers3drototranslationequivariant}, five particles each carry either a positive or a negative charge and exert repulsive or attractive forces on each other. The network input is the position of a particle in a specific time step, its velocity, and its charge. The algorithm must predict the relative location and velocity 500 time steps into the future.

For this test, our model consists of a equivariant graph convolution~\citep{brandstetter2022geometricphysicalquantitiesimprove}, followed by 4 equivariant Clebsch-Gordan attention layers. We use irriducibles types up to six, each with channel dimension of 8 and head dimension of four. Models were trained for 500 epochs, using cosine annealing scheduler with an initial learning rate of $1e^{-3}$ and gradient clipping. Each run was run on a single NVIDA V100 GPU.

\begin{figure}
\centering
\includegraphics[width=0.8\textwidth]{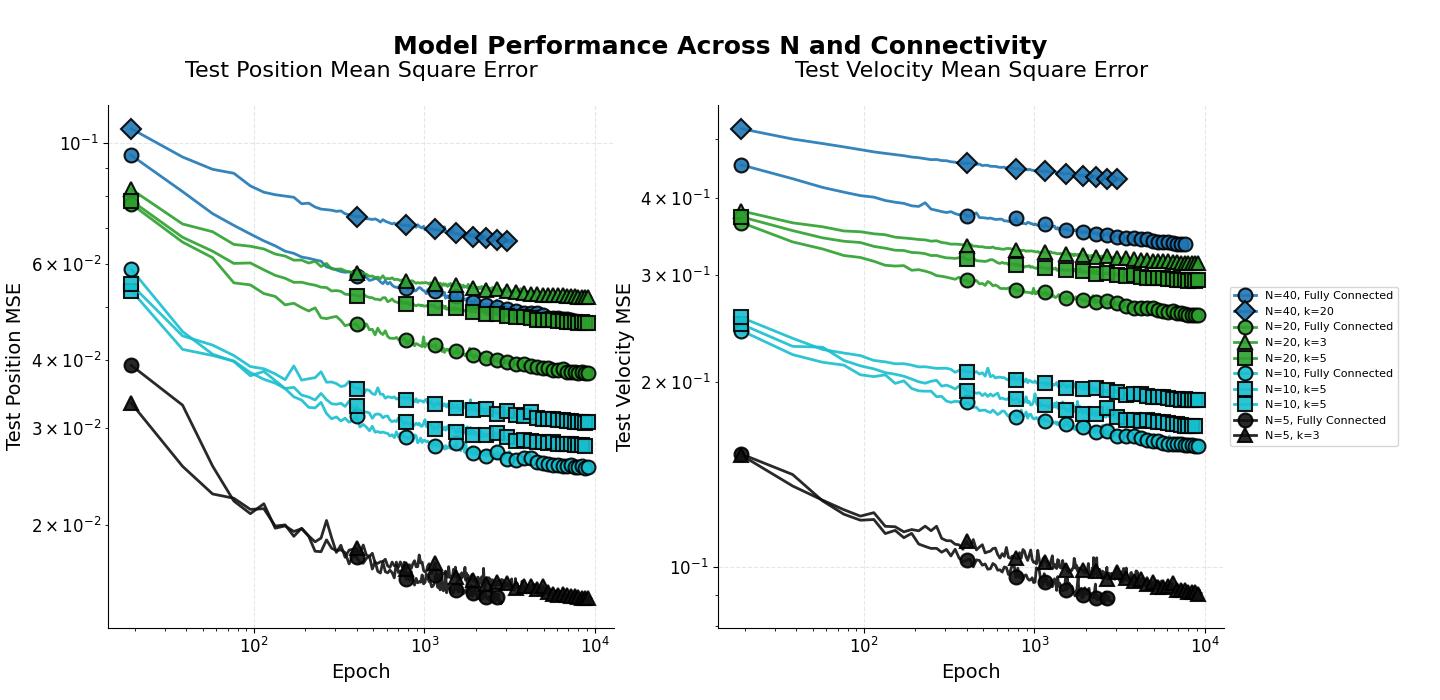}
\caption{Performance of $SE(3)$-transformer on the Nbody dataset for different number of points $N$ and nearest neighbors $k$. We use the exact model parameters at that of \citet{fuchs2020se3transformers3drototranslationequivariant}. Note that on the $N=20$ curves decreasing $k=20$ (fully connected graph) to $k=5$ leads to over a twenty percent drop in performance. Left shows MSE of position, right shows MSE of velocity. Local context window based methods fail to capture accuracy as tasks get more difficult; global context is needed.}
\label{fig:Performance Profile}
\end{figure}

As motivation, we show that the model performance can vary with the size of the message passing window. Specifically, in the nbody simulations of \citet{fuchs2020se3transformers3drototranslationequivariant,brandstetter2022geometricphysicalquantitiesimprove,satorras2022enequivariantgraphneural} an all to all connection is used. This will scale quadratically in the number of particles, which quickly becomes computationally untractable. We \ref{fig:Performance Profile} perform an ablation study on \citet{fuchs2020se3transformers3drototranslationequivariant} and \citet{brandstetter2022geometricphysicalquantitiesimprove} where we use a $k$-nearest neighbors, as opposed to fully connected graphs.

\begin{table}[H]
\centering
\resizebox{\textwidth}{!}{%
\begin{tabular}{|l|cc|ccc|cccc|ccccc|}
\hline
\multirow{2}{*}{\textbf{Method}}
& \multicolumn{2}{c|}{\textbf{N = 5}}
& \multicolumn{3}{c|}{\textbf{N = 10}}
& \multicolumn{4}{c|}{\textbf{N = 20}}
& \multicolumn{5}{c|}{\textbf{N = 40}} \\
\cline{2-15} 
& $k$=3 & fc & $k$=3 & $k$=5 & fc & $k$=3 & $k$=5 & $k$=10 & fc & $k$=3 & $k$=5 & $k$=10 & $k$=20 & fc \\
\hline
SE(3)-Transformer & 0.013 & 0.013 & 0.031 & 0.028 & 0.025 & 0.057 & 0.052 & 0.050 & 0.044 & 0.061 & 0.056 & 0.052 & 0.049 & OOM \\
\hline
SEGNN & 0.040 & 0.048 & 0.023 & 0.018 & 0.013 & 0.042 & 0.039 & 0.033 & 0.029 & 0.052 & 0.480 & 0.450 & 0.038 & OOM \\
\hline
Ours w/o local &  & \textbf{0.003} &  &  & 0.012 &  &  &  & 0.026 &  &  &  &  & 0.031 \\
\hline
Ours &  & \textbf{0.003} &  & & \textbf{0.010} &  &  &  & \textbf{0.023} &  &  &  &  & \textbf{0.030} \\
\hline
\end{tabular}%
}
\caption{Performance comparison for varying numbers of particles $N$ and nearest neighbors $k$ (fc = fully connected graph). Our method uses fixed $k=3$ local attention. Averaged over 2 seeds.}
\label{tab:nbody-comp}
\end{table}

As is shown in \cref{tab:nbody-comp}, our method achieves state of the art performance on this task. Furthermore, our model is highly scalable to larger $N$. Although this is a toy task, we believe these results illustrates both the scalability of our method and the danger of using methods with only local context. Additional experiments and ablations are in \cref{Appendix: Additional Experiments}.

\subsection{QM9}

For molecular chemistry, we benchmark on the QM9 dataset \citep{ramakrishnan2014quantum}, a widely-used collection of 134k small organic molecules with up to 9 heavy atoms (C, O, N, F). Each molecule is annotated with 19 regression targets, including atomization energies, dipole moments, and HOMO-LUMO gaps, calculated using DFT. Following standard practice, we predict one target at a time using the provided 110k/10k/10k training/validation/test split, and report mean absolute error (MAE) in units consistent with prior work.


In its current form, our model does not use edge features. For that reason, we first apply an SEGNN layer~\citep{brandstetter2022geometricphysicalquantitiesimprove} with skip connection, followed by  four of our attention blocks. Output invariant features are then fed into an MLP for classification. For this test, our model consists of an equivariant graph convolution~\citep{brandstetter2022geometricphysicalquantitiesimprove}, followed by 8 equivariant Clebsch-Gordan attention layers. We use irreducibles up to $\ell=6$, each with 8 channels and 4 heads. Models were trained for 500 epochs, using cosine annealing scheduler with an initial learning rate of $1e^{-4}$ and gradient clipping. Each run was run on eight NVIDA V100 GPUs. Additional experiments and ablations are shown in \cref{Appendix: Additional Experiments}.

\begin{table*}[t!]
\centering
\small
\begin{tabular}{|l|c|c|c|c|c|c|}
\hline
\multirow{3}{*}{\textbf{Model}} & \multicolumn{6}{c|}{\textbf{Mean Absolute Error}} \\
\cline{2-7}
& $\alpha$ & $\Delta\epsilon$ & $\epsilon_{\text{HOMO}}$ & $\epsilon_{\text{LUMO}}$ & $\mu$ & $C_v$ \\
& (Bohr$^3$) & (meV) & (meV) & (meV) & (D) & (cal/mol$\cdot$K) \\
\hline
SE(3)-Transformer & 0.142 & 53.0 & 35.0 & 33.0 & 0.510 & 0.054 \\
\hline
EGNN & 0.071 & 48.0 & 29.0 & 25.0 & 0.290 & 0.031 \\
\hline
SEGNN & 0.060 & 42.0 & 24.0 & 21.0 & 0.023 & 0.310 \\
\hline
Ours w/o local & 0.100 & 49.0 & 31.0 & 26.0 & 0.310 & 0.350 \\
\hline
\textbf{Ours} & 0.100 & \textbf{39.0} & \textbf{26.0} & \textbf{19.0} & \textbf{0.210} & \textbf{0.030} \\
\hline
\end{tabular}
\caption{Mean absolute error (MAE) on QM9 dataset. Lower values indicate better performance.}
\label{tab:qm9_results}
\end{table*}

\subsection{ModelNet40 Classification}

The ModelNet40 classification task is a widely used benchmark for evaluating 3D shape recognition methods. Introduced by \citet{wu20153d}, the ModelNet40 dataset consists of $12,311$ 3D CAD models from $40$ object categories. The dataset is split into $9,843$ training examples and $2,468$ test examples. The task is to classify 3D objects based on geometric structure. We compare our model with DGCNN~\citep{wang2019dynamicgraphcnnlearning} and PointNet++~\citep{qi2017pointnetdeephierarchicalfeature} which are state of the art non-equivariant methods.

\begin{figure}[H]
\begin{minipage}[t]{0.48\textwidth}
\centering
\vspace{0pt}
\label{tab:modelnet-results}
\vspace{2pt}
{\footnotesize
\begin{tabular}{@{}lccc@{}}
\toprule
\textbf{Model} & \textbf{Year} & \textbf{MN10} & \textbf{MN40} \\
& & \textbf{Acc.(\%)} & \textbf{Acc.(\%)} \\
\midrule
DGCNN & 2019 & 95.1 & 92.9 \\
PointNet++ & 2020 & 97.4 & 91.3 \\
SEGNN & 2021 & 94.2 & 90.5 \\
SE(3)-Trans. & 2020 & 93.2 & 88.1 \\
\midrule
Ours w/o local & 2025 & 95.5 & 89.3 \\
Ours & 2025 & 90.1 & 85.9 \\
\bottomrule
\end{tabular}
\captionof{table}{Classification accuracy on ModelNet10 and ModelNet40. All models trained with scale, rotation, and permutation augmentation.}
}
\end{minipage}%
\hspace{-10pt}
\begin{minipage}[t]{0.48\textwidth}
\centering
\vspace{0pt}
\includegraphics[width=0.525\linewidth]{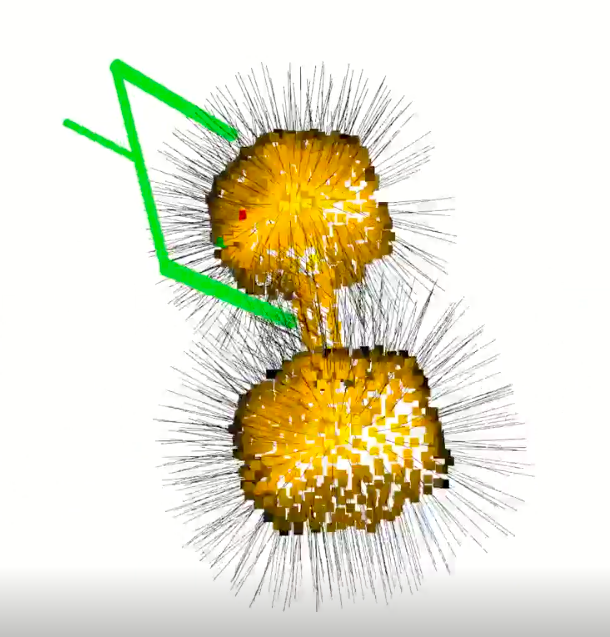} 
\captionof{figure}{Robotic object dataset example showing an object with surface normals and optimal grasp locations.}
\label{fig:robotic-grasp}
\end{minipage}
\vspace{-20pt}
\end{figure}

\subsection{Object Grasping Dataset}
We also consider a bespoke robotic grasping dataset. Robotic grasping fundamentally depends on object geometry, and high precision robotic grasping is difficult because it requires both fine angular resolution and large context window. Our dataset consists of 400 samples, each of which consists of a point cloud, a set of surface normal vectors, an optimal grasp orientation (represented as a $3 \times 3$ matrix), a optimal grasp depth (which is a single positive real number) and an optimal grasp location. Each point cloud has resolutions of 512, 1024, 2048, or 4096 points. We consider three tasks for the object grasping dataset; namely, surface normal prediction and grasp prediction. We detail these tasks in \ref{Section:Object Grasping Dataset}.

\begin{table}[ht]
\centering
\resizebox{\textwidth}{!}{%
\begin{tabular}{|l|c|cccc|cccc|cccc|cccc|}
\hline
\multirow{2}{*}{\textbf{Model}} & \multirow{2}{*}{\textbf{Year}} &
\multicolumn{4}{c|}{\textbf{Rotation Error}} &
\multicolumn{4}{c|}{\textbf{Distance Error}} &
\multicolumn{4}{c|}{\textbf{Depth Error}} &
\multicolumn{4}{c|}{\textbf{Normal Error}} \\
\cline{3-18}
& & 512 & 1024 & 2048 & 4096 & 512 & 1024 & 2048 & 4096 & 512 & 1024 & 2048 & 4096 & 512 & 1024 & 2048 & 4096 \\
\hline
DGCNN & 2018 & 0.015 & 0.019 & 0.031 & 0.120 & 3.01 & 5.34 & 8.34 & 10.30 & 0.08 & 0.08 & 0.09 & 0.08 & 0.013 & 0.017 & 0.023 & 0.051 \\
\hline
PointNet++ & 2017 & 0.015 & 0.021 & 0.028 & 0.080 & 2.51 & 4.88 & 5.57 & 9.54 & 0.08 & 0.08 & 0.07 & 0.09 & 0.013 & 0.018 & 0.021 & 0.048 \\
\hline
SEGNN & 2021 & 0.018 & 0.024 & 0.031 & 0.100 & 4.02 & 5.26 & 8.37 & 10.54 & 0.08 & 0.09 & 0.08 & 0.09 & 0.015 & 0.023 & 0.025 & 0.052 \\
\hline
SE(3)-Trans. & 2020 & 0.025 & 0.028 & OOM & OOM & 5.03 & 7.91 & OOM & OOM & 0.08 & 0.09 & OOM & OOM & 0.025 & 0.035 & OOM & OOM \\
\hline
Ours w/o local & 2025 & 0.019 & 0.025 & 0.030 & 0.090 & 4.02 & 5.10 & 8.30 & 9.85 & 0.08 & 0.08 & 0.08 & 0.08 & 0.013 & 0.017 & 0.020 & 0.041 \\
\hline
\textbf{Ours} & 2025 & \textbf{0.013} & \textbf{0.017} & 0.025 & \textbf{0.080} & \textbf{2.44} & \textbf{3.51} & \textbf{5.31} & \textbf{9.39} & 0.08 & 0.08 & \textbf{0.07} & 0.08 & \textbf{0.011} & \textbf{0.015} & \textbf{0.019} & \textbf{0.039} \\
\hline
\end{tabular}%
}
\caption{Performance comparison on robotic grasping dataset across different point cloud resolutions. OOM = Out of Memory. All models trained with rotation, permutation, and scaling augmentation.}
\label{tab:grasping-results}
\end{table}

We benchmarked each of the baselines methods using the same model parameters as model net classification. Harmonics and nearest neighbors were chosen to be the max amount that fit on memory. Each training run was done on 8 NVIDIA v100 GPUs for 500 epochs.

\section{Conclusion}



\textbf{Conclusions.} 
This work tackles two key challenges in equivariant transformers: achieving scalability to global geometric context and efficiently computing high-order irreducible representations for greater expressiveness. By extending vector long convolution to Clebsch–Gordon convolution, we propose the first architecture that achieves global token attention in $\mathcal{O}(N \log N)$ time and supports arbitrary orders of irreducible representations. In addition, our method allows for permutation equivariance, which is essential for point cloud and atomic systems. We also provide comprehensive theoretical analyses to prove both equivariance and time complexity. Finally, we benchmark our method on various dataset, demonstrating clear gains in memory, speed, and accuracy across robotics, physics, and chemistry, outperforming existing state-of-the-art equivariant transformers.

\textbf{Limitations.}
The $\mathcal{O}(L^{3})$ complexity of our method is still suboptimal and may be prohibitive for cases requiring very high angular resolution.  

\textbf{Future work.}
Equivariant transformers can draw significant inspiration from computational astrophysics methods designed to efficiently handle $N$-body interactions. 
\bibliography{refs}

\newpage

\appendix
\section{Additional Experiments}\label{Appendix: Additional Experiments}

The goal of this paper is to develop a model architecture that is optimal in harmonic and number of tokens and can learn good representations across a variety of tasks. We summarize our benchmarking results in \ref{tab:baseline-benchmarks}.

\begin{table}[h]
\centering
\caption{Summary of benchmarking results across datasets. Black checkmarks = our reproduced results, \textcolor{blue}{blue checkmarks} = results from original papers.}
\label{tab:baseline-benchmarks}
\vspace{2pt}
\begin{tabular}{@{}lcccccc@{}}
\toprule
\textbf{Method} & \textbf{Year} & \textbf{QM9} & \textbf{ModelNet40} & \textbf{Robotic} & \textbf{N-body} \\
& & & \textbf{Classification} & \textbf{Grasp} & \\
\midrule
\multicolumn{6}{l}{\textit{Equivariant Methods}} \\
SEGNN & 2021 & \textcolor{blue}{\checkmark} & \checkmark & \checkmark & \checkmark \textcolor{blue}{\checkmark} \\
SE(3)-Transformer & 2020 & \checkmark \textcolor{blue}{\checkmark} & \checkmark & \checkmark & \checkmark \textcolor{blue}{\checkmark} \\
SE(3)-Hyena$^*$ & 2025 & --- & \checkmark & \checkmark & \checkmark \textcolor{blue}{\checkmark} \\
Equiformer-V2 & 2023 & \textcolor{blue}{\checkmark} & --- & \checkmark & \textcolor{blue}{\checkmark} \\
GATr & 2025 & \textcolor{blue}{\checkmark} & --- & --- & \checkmark \textcolor{blue}{\checkmark} \\
\midrule
\multicolumn{6}{l}{\textit{Non-Equivariant Methods}} \\
DGCNN & 2018 & --- & \checkmark \textcolor{blue}{\checkmark} & \checkmark & \checkmark \textcolor{blue}{\checkmark} \\
PointNet++ & 2017 & --- & \checkmark \textcolor{blue}{\checkmark} & --- & --- \\
\bottomrule
\end{tabular}
\begin{tablenotes}
\footnotesize
\item $^*$Our implementation
\end{tablenotes}
\end{table}

\section{Object Grasping Dataset}\label{Section:Object Grasping Dataset}
As mentioned in the main text, we benchmark on a custom robotic grasping dataset. Robotic grasping fundamentally depends on object geometry, and high precision robotic grasping is a difficult problem because it requires both fine angular resolution and large context window. Our dataset consists of 400 samples, each of which consists of a point cloud, a set of surface normal vectors, an optimal grasp orientation (represented as a $3 \times 3$ matrix), a optimal grasp depth (which is a single positive real number) and an optimal grasp location (i.e. the index of one point in the point cloud). Each group of 100 samples has point cloud resolutions of 512, 1024, 2048, and 4096 points, respectively. Our dataset will be made publicly available.

.\subsubsection{Surface Normal Prediction}

In the first task, we the model is given the point cloud and must predict the surface normals. This task has inherent $SO(3)$-symmetry as the normal vectors are rotated if the point cloud is rotated. We use a $80-10$ split for train and test data. Each model is given $N$ input vectors (type 1 features) and must output $N$ vectors (type 1) features. We benchmarked each of the baselines methods using the same model parameters as model net classification. Harmonics and nearest neighbors were chosen to be the max amount that fit on memory. Each training run was done on 8 NVIDIA v100 GPUs. Each model was trained for 500 epochs.

\subsubsection{Grasping}
To better simulate real-world settings, we begin by sampling an object and dropping it onto a table. We then randomly select 
$n$ camera views to capture depth images. These depth images are projected into 3D space using the corresponding camera matrices to generate point clouds. We segment out the table from each raw point cloud and apply the Farthest Point Sampling (FPS) algorithm to downsample the remaining points to the desired resolution, ensuring no duplicate points. To obtain the optimal grasp pose, we first sample an approach point, an orientation, and a grasp depth. We then execute the grasp and evaluate its success by checking whether the object can be lifted and remains secure during a sequence of gripper shaking motions.

\subsubsection{Grasp Prediction}
In the grasp prediction task, the model is given the point cloud and the surface normals and must predict a distribution for optimal grasp location, an optimal grasp orientation and an optimal grasp depth. It should be noted that many of these objects have symmetries, and can have multiple optimal grasp locations. We define the weighted distance error as
\begin{align*}
D( p_{true} , p_{model} ) = \int drdr' \text{ } p_{true}(r)d(r,r')p_{model}(r') 
\end{align*}
which measure the optimal transport distance between the true distribution $p_{true}$ and the model output $p_{model}$. Grasping is an inherently difficult task as finding the optimal grasp location requires processing information about the global properties of the the object. For this reason, we expect that local attention methods will have difficulty with this task.

\section{Adding Permutation Equivariance}

One limitation of~\citet{moskalev2024se3hyenaoperatorscalableequivariant} is that the Fourier decomposition breaks permutation equivariance. Specifically, the introduction of the choice of ordering to compute the Fourier basis breaks that natural permutation equivariance of points. We say that a transformer is permutation equivariant if under the input transformation $u_{i} \rightarrow u_{\sigma(i)}$ the queries, keys and values are transformed via the permutation
\begin{align*}
q_{i} \rightarrow q_{\sigma(i)}, \quad k_{i} \rightarrow k_{\sigma(i)}, \quad v_{i} \rightarrow v_{\sigma(i)}
\end{align*}
so that the attention matrix transforms as 
\begin{align*}
\alpha_{ij} \rightarrow \alpha_{ \sigma(i) \sigma(j) }
\end{align*}
Transformers without positional embeddings are naturally permutation invariant, and permuting inputs results in a different permutation of outputs. 
Methods like deep set transformer~\cite{lee2019settransformerframeworkattentionbased} enforce permutation invariance by adding additional weight tying constraints. The method proposed in~\cite{moskalev2024se3hyenaoperatorscalableequivariant} is not permutation invariant as it relies on Hyena attention~\cite{poli2023hyenahierarchylargerconvolutional}, which is itself not permutation equivariant.  Transformer permutation equivariance is key for point cloud operations and simulation of atomic systems (which are, by the uncertainty principle, indistinguishable). Models which are not hardcoded to be permutation invariant will not be able to automatically generalize to permutations of the input graph. We tested numerous methods for enforcing permutation equivariance, discussed in \cref{Appendix: Model Ablations}. In general, we found that data augmentation with regularization or Fourier space attention was optimal. We describe the Fourier space attention in this section, and two other methods, along with ablation study, are discussed in \cref{Appendix: Model Ablations}.

\subsubsection{Fourier Space Attention}

The Fourier transform on graphs extends classical Fourier analysis to signals defined on the vertices of a graph. Let $G = (V, E)$ be an undirected graph with $n$ nodes, and let $L \in \mathbb{R}^{n \times n}$ denote its (combinatorial) normalized graph Laplacian, defined as $L = I - D^{-\frac{1}{2}} AD^{-\frac{1}{2}}$, where $A$ is the adjacency matrix, $D$ is the degree matrix and $I$ is the identity matrix. The graph Laplacian $L$ is symmetric and positive semi-definite, it admits an eigendecomposition $L = U \Lambda U^\top$ where $U = [u_1, \ldots, u_n]$ is an orthonormal matrix of eigenvectors and $\Lambda = \operatorname{diag}(\lambda_1, \ldots, \lambda_n)$ is the diagonal matrix of positive eigenvalues. The eigenvectors of $L$ form the basis for the graph Fourier transform. For a signal $x \in \mathbb{R}^{n \times d}$ defined on the graph's nodes, its GFT is given by $\hat{x} = U^\top x$ with the inverse GFT is $x = U \hat{x}$. Intuitively, the eigenvectors of the Laplacian serve as generalized ``Fourier modes'' on the graph, and the eigenvalues correspond to their frequencies. This spectral perspective enables convolution and filtering operations on graphs by manipulating the signal in the frequency domain. Let \( x \in \mathbb{R}^{n \times d'} \) be a signal on the graph and \( g \in \mathbb{R}^{ n \times d \times d' } \) be a filter. The graph convolution in frequency space is defined as:
\[
x \star_G g = U \left( (U^\top g) \odot (U^\top x) \right) = U ( \hat{g} \odot \hat{x} )
\]
where \( \hat{x} = U^\top x \), \( \hat{g} = U^\top g \) are the signal and filter Fourier transforms, and \( \odot \) denotes element-wise multiplication. This operation corresponds to pointwise multiplication in the spectral domain followed by an inverse transform. The convolution in Fourier space is permutation equivariant. To see this, note that under a permutation of inputs, the Laplacian matrix transforms as $ L \rightarrow U_{\sigma} L U_{\sigma}^{-1}$ where $U_{\sigma}$ denotes the matrix that is a permutation of inputs. Thus, the diagonalization of $L$ transforms as
\begin{align*}
L = U \Lambda U^{-1} \rightarrow U_{\sigma} L U_{\sigma}^{-1} = U_{\sigma}[ U \Lambda U^{-1}  ]U_{\sigma}^{-1} = [U_{\sigma}U] \Lambda [ U^{-1} U_{\sigma}^{-1}]
\end{align*}
so the matrix $U$ transforms as $U\rightarrow U_{\sigma} U$. Thus, under rank spectral convolution, 
\begin{align*}
x \star_G g = U \left( (U^\top g) \odot (U^\top x) \right) \rightarrow U_{\sigma} U \left( (U^\top g) \odot (U^\top x) \right) = U_{\sigma} ( x \star_G g )
\end{align*}
so the spectral convolution $x \star_G g$ is permutation equivariant.





\section{Model Baselines}\label{Appendix: Model Baselines}
In this section, we describe the model baselines in more depth. We split models into two categories, equivariant and non-equivariant. For fairness, both model categories were always trained using data augmentation. 

\subsection{Non-Equivariant Models}

For the non-equivariant models, we use DGCNN \cite{wang2019dynamicgraphcnnlearning} and PointNet++ \cite{qi2017pointnetdeephierarchicalfeature}.

\emph{DGCNN} (Dynamic Graph CNN) constructs local neighborhood graphs in each layer and applies convolution-like operations on edges, allowing the model to capture local geometric structures. DGCNN only utilizes local operations, and may have issues modeling long range dependencies. 

\emph{PointNet++} extends the original PointNet by applying hierarchical feature learning on nested partitions of the input point set, capturing both local and global context in point clouds. PointNet++ specifically uses layers designed to capture information across multiple length scales, which may explain its excellent performance, despite not being equivariant.

\subsection{Equivariant Models}

For the equivariant models, we benchmark against Tensor Field Networks \cite{thomas2018tensorfieldnetworksrotation}, SE3-transformer \cite{fuchs2020se3transformers3drototranslationequivariant}, SEGNN \cite{brandstetter2022geometricphysicalquantitiesimprove} and SE3-Hyena \cite{moskalev2024se3hyenaoperatorscalableequivariant}.

\emph{TFN}  
Tensor Field Networks (TFNs)~\cite{thomas2018tensorfieldnetworksrotation} are 3D rotation and translation equivariant networks that associate each point with features transforming under irreducible representations of $SO(3)$. Using spherical harmonics and learnable radial functions, TFNs perform equivariant convolutions over point clouds or molecules.

\emph{SE3-Transformer}  
The SE(3)-Transformer~\cite{fuchs2020se3transformers3drototranslationequivariant} builds on the tensor field network for point clouds with SE(3)-equivariance. It uses self-attention and tensor-based operations to model interactions in 3D space, making it effective for tasks like molecular modeling.

\emph{SEGNN}  
SEGNN~\cite{brandstetter2022geometricphysicalquantitiesimprove} is a message passing neural network that ensures SE(3)-equivariance by representing features with $\mathrm{SO}(3)$ irreducibles. SEGNN is designed to model physical quantities that transform under 3D symmetries. Another contribution of \cite{brandstetter2022geometricphysicalquantitiesimprove} was to formalize the connection between message passing methods and transformers.

\emph{SE(3)-Hyena}  
SE(3)-Hyena~\cite{moskalev2024se3hyenaoperatorscalableequivariant} is the most similar method we compare to. It introduces SE(3)-equivariant Hyena operators that combine hierarchical long-range interactions with invariant and vector representations.

\subsection{Tensor Product Ablation}\label{Appendix: Tensor Product Ablation }

We explain in more depth the tensor product computation.

\subsubsection{Related Work}

The tensor product non-linearity requires significant memory usage. Numerous methods have been developed to reduce the computational complexity of the tensor product. The Gaunt tensor product method \cite{liaoequiformer} changes the Clebsch-Gordan tensor product to the overlap of three spherical harmonics, resulting in an $\mathcal{O}(L^{3})$ complexity. However, as pointed out in \cite{xie2024price}, the Gaunt tensor product is unable to distinguish parity, which makes it unsuitable for many molecular data processing applications where distinguishing compounds of different chirality is essential (i.e. thalidomide disaster). A different line of work attempts to speed up the tensor product operation using the fast fourier transform on the sphere. The Fourier transform on groups relates $SO(3)$ features to functions on the sphere. The tensor product of $SO(3)$ features is directly related to their product in real space. However, spherical Fourier transforms often suffer from numerical accuracy issues, and this problem is exacerbated on GPU.

\begin{table}[h]
\centering
\caption{Comparison of Tensor Product Methods}
\begin{tabular}{lccc}
\toprule
\textbf{Method} & \textbf{Year} & \textbf{Scaling} & \textbf{GPU Parallelized?} \\
\midrule
Naive Method &  & $\mathcal{O}(L^{6})$ & yes \\
Sparse Method &  & $\mathcal{O}(L^{3})$ & yes \\
Gaunt Tensor Product & 2024 & $\mathcal{O}(L^{3})$ & yes \\
\midrule
Fourier Transform (HEALPix) & 2022 & $\mathcal{O}(L^{3})$ & no \\
Fourier Transform (ssht) & 2011 & $\mathcal{O}(L^{3})$ & yes \\
Fourier Transform (s2fft) & 2024 & $\mathcal{O}(L^{3})$ & yes \\
Fourier Transform \cite{driscoll=healy} & 1995 & $\mathcal{O}(L^{2} \log^{2}(L) )$ & no \\
Fourier Transform \cite{Suda2002AFS} & 2002 & $\mathcal{O}(L^{2} \log L)$ & no \\
\bottomrule
\end{tabular}
\label{tab:tensor-product-comparison}
\caption{Summary of Tensor product methods. The methods based on Fourier transformation have some issues with numerical accuracy and large constant memory cost, and may be suboptimal.  }
\end{table}

\subsubsection{Problem Statement}
Let $\{ f^{\ell} \}_{\ell=0}^{L}$ and $\{ g^{\ell} \}_{\ell=0}^{L}$ be a set of input features band-limited (of degree $\ell$) at $L$. We wish to compute the full tensor product
\begin{align*}
h^{J} = \sum_{\ell} C^{J}_{\ell \ell'} f^{\ell} \otimes g^{\ell'}
\end{align*}
for all $J$ in the range $0,1,2,...,L_{out}$, where $L_{out}$ is the maximum output harmonic. Features have additional head and channel dimensions. Separate heads do not interact. If two input features have channel dimension $m$ and $m'$, the tensor product of these features has channel dimension $m \cdot m'$. Now, using the tensor product decomposition
\begin{align*}
\ell \otimes \ell' = \bigoplus_{k=|\ell - \ell'|}^{\ell + \ell'} k
\end{align*}
the naive complexity of computing this tensor product is given by computing every term in the expansion of $h^{J}$. The total computational complexity of computing the single term $C^{J}_{\ell \ell} f^{\ell} \otimes g^{\ell'} $ involves contraction of a tensor of size $(2J+1)(2\ell+1)(2\ell'+1)$ and tensors of size $(2\ell+1)$ and $(2\ell'+1)$ for a total complexity of $\mathcal{O}( (2J+1)(2\ell+1)(2\ell'+1) )$. Thus, the total complexity of computing each term in the tensor product is
\begin{align*}
h^{0}: \sum_{ij=1,|i-j|=1 }^{L} O( 1ij )  &= \mathcal{O}(L^{3}) \\
h^{1}: \sum_{ij=1,|i-j|=2 }^{L} O( 3ij ) &= \mathcal{O}(L^{3}) \\
&\vdots \\
h^{L}:\sum_{ij=1,|i-j|=L }^{L} O( (2L+1)ij ) &= \mathcal{O}(L^{4})
\end{align*}
Thus, the total computational complexity of computing all the resultant terms $h^{J}$ is $O( L^{6} )$.

\subsubsection{Sparsity}
However, the fact that the matrix $C^{J}_{\ell \ell'}$ is sparse can be used to reduce the computational overhead. Specifically, in the real basis, harmonics are zero only when $M = \pm m + \pm' m'$. Thus, the matrix multiplication
\begin{align*}
h^{J} = C^{J}_{\ell \ell} f_{\ell} \otimes g_{\ell'}
\end{align*}
can be performed by computing the non-zero elements of $C^{J}_{\ell \ell'}$. There are $(2\ell +1)(2\ell'+1)$ such non-zero matrix entries. Thus, the computational complexity in computing the full tensor product is
\begin{align*}
h^{}: \sum_{ij=1,|i-j|=1 }^{L} O( ij )  &= \mathcal{O}(L^{3}) \\
h^{1}: \sum_{ij=1,|i-j|=2 }^{L} O( ij ) &= \mathcal{O}(L^{3}) \\
&\vdots \\
h^{L}:\sum_{ij=1,|i-j|=L }^{L} O( ij ) &= \mathcal{O}(L^{3})
\end{align*}
so that the total computational complexity of computing each $h^{J}$ is $O( L^{5} )$.

\subsubsection{Fourier Transform Method}

The Fourier coefficients specify a function on the sphere. Specifically, any function $F: S^{2} \rightarrow \mathbb{C}$ can be written as
\begin{align*}
F( \hat{n} ) = \sum_{\ell=0}^{\infty} ( F^{\ell} )^{T} Y^{\ell}( \hat{n} )
\end{align*}
where the Fourier transform coefficients are given by
\begin{align*}
F^{\ell} = \int_{ \hat{n} \in S^{2} } d\hat{n} \enspace F(\hat{n})Y^{\ell}(\hat{n})
\end{align*}
The tensor product of the Fourier coefficients corresponds to multiplication in Fourier space. Specifically, let $f$ and $g$ be functions on the sphere with Fourier expansions
\begin{align*}
f( \hat{n} ) = \sum_{\ell=0}^{\infty} (f^{\ell})^{T}Y^{\ell}( \hat{n} ), \quad g( \hat{n} ) = \sum_{\ell=0}^{\infty} (g^{\ell})^{T}Y^{\ell}( \hat{n} ) 
\end{align*}
then, consider the Fourier expansion of the product $fg = f \cdot g$ given by
\begin{align*}
(fg)(\hat{n}) = f( \hat{n} )g( \hat{n} ) = \sum [ (fg)^{\ell} ]^{T} Y^{\ell}(\hat{n})
\end{align*}
Using the inverse Fourier transform in the sphere
\begin{align*}
(fg)^{\ell} = \int_{ \hat{n} \in S^{2} } d\hat{n} \enspace (fg)(\hat{n} ) Y^{\ell}(\hat{n})
\end{align*}
Now, using the Fourier expansions of $f$ and $g$, we have that
\begin{align*}
(fg)(\hat{n}) &= \left( \sum_{\ell_1=0}^{\infty} [f^{\ell_1}]^{T} Y^{\ell_1}(\hat{n}) \right) \left( \sum_{\ell_2=0}^{\infty} [ g^{\ell_2} ]^{T} Y^{\ell_2}(\hat{n}) \right) = \sum_{\ell_1, \ell_2=0}^{\infty} [f^{\ell_1} \otimes g^{\ell_2} ]^{T} [ Y^{\ell_1}(\hat{n}) \otimes Y^{\ell_2}(\hat{n}) ] \\
\end{align*}
The tensor product of two spherical harmonics decomposes as
\begin{align*}
Y^{\ell_1}(\hat{n}) \otimes Y^{\ell_2}(\hat{n}) =
\sum_{\ell = |\ell_1 - \ell_2|}^{\ell_1 + \ell_2}
C^{\ell}_{\ell_1 \ell_2} \, Y^{\ell}(\hat{n})
\end{align*}
where $C^{\ell_1, \ell_2}_{\ell}$ are the Clebsch-Gordan coefficients. Thus, we have that
\begin{align*}
(fg)(\hat{n}) &= \sum_{\ell_1, \ell_2=0}^{\infty} [f^{\ell_1} \otimes g^{\ell_2} ]^{T} [ Y^{\ell_1}(\hat{n}) \otimes Y^{\ell_2}(\hat{n}) ] = \sum_{\ell_1, \ell_2=0}^{\infty} \sum_{\ell = |\ell_1 - \ell_2|}^{\ell_1 + \ell_2} [f^{\ell_1} \otimes g^{\ell_2} ]^{T} 
C^{\ell}_{\ell_1 \ell_2} \, Y^{\ell}(\hat{n}) \\
\end{align*}
and reindexing this sum,
\begin{align*}
(fg)(\hat{n}) = \sum_{\ell_1, \ell_2=0}^{\infty} \sum_{\ell = |\ell_1 - \ell_2|}^{\ell_1 + \ell_2} [f^{\ell_1} \otimes g^{\ell_2} ]^{T} 
C^{\ell}_{\ell_1 \ell_2} \, Y^{\ell}(\hat{n}) = \sum_{\ell=0}^{\infty} [ \sum_{\ell_1,\ell_2=0}^{\infty} C^{\ell_1 \ell_2}_{\ell}( f^{\ell_1} \otimes g^{\ell_2} ) ]^{T} 
 Y^{\ell}(\hat{n})
\end{align*}
Therefore, the Fourier coefficients of the product are given by
\begin{align*}
[ (fg)^{\ell} ]^{T} = \sum_{\ell_1, \ell_2} C_{\ell}^{\ell_1, \ell_2} [ f^{\ell_1} \otimes g^{\ell_2} ]^{T}
\end{align*}
Thus, one way to evaluate the tensor product is to compute the inverse spherical harmonic transform (isht) of $f^{\ell}$ and $g^{\ell}$, multiply the signals on the sphere, and then spherical Fourier transform (sht). We consider a few methods for computing the sht and isht.

In principal, the computational complexity of computing the Fourier transform can be as low as $\mathcal{O}(L^{2} \log L )$. However, for the regime of interest ($L \approx 7$), methods that have $\mathcal{O}(L^{3})$ complexity often have lower prefactor and are more memory efficient.

\emph{Healpix}
HEALPix (Hierarchical Equal Area isoLatitude Pixelization) ~\cite{Gorski_2005} is a standard method for discretizing functions on the sphere. HEALPix sampling divides the sphere into equal-area pixels arranged along iso-latitude rings. HEALPix has built in support for spherical harmonic transforms. The full cost of spherical harmonic transform on HEALpix grid is $\mathcal{O}(L^{3})$.

\emph{ssht}
ssht \cite{mcewen:fssht} library is a C++ and Python library for computing spherical harmonic transforms. It is used for Fourier analysis on the sphere, particularly in scientific fields like astrophysics and geophysics. ssht \cite{mcewen:fssht} uses a novel sampling scheme on the sphere to compute the Fourier transform. The computationally complexity of this method scales as $\mathcal{O}(L^3)$.

\emph{s2fft}
s2fft \cite{Price_2024} is a recursive algorithm for the calculation of Wigner d-functions. s2fft is a highly parallelizable method and is implemented in JAX. s2fft supports HEALpix and equiangular sampling. The computationally complexity of this method scales as $\mathcal{O}(L^3)$, but by utilizing GPU parallelization, the complexity decreases almost linearly in number of GPUs used. 

\cite{cobb2021efficientgeneralizedsphericalcnns} also developed interesting methods for spherical convolutions which may be utilized for fast group theoretic Fourier transformations in the future.

\subsection{Tensor Product Ablations}

We compared three methods for computing the tensor product, the naisve method, the gaunt tensor product method, the Fourier transform method with healpix, the Fourier transform with s2fft\cite{Price_2024} and ssht \cite{mcewen:fssht} and the sparse method.

\begin{table*}[ht]
\centering
\caption{Mean absolute error (MAE) on QM9 dataset using different tensor product methods. Lower is better.}
\label{tab:qm9_tensor_product_methods}
\setlength{\tabcolsep}{3pt}
\fontsize{8pt}{3pt}\selectfont

\resizebox{\textwidth}{!}{
\begin{tabular}{|l|c|c|c|c|c|c|}
\hline
\textbf{Tensor Product Method} &  $\alpha$ (Bohr$^3$) & $\Delta \epsilon$ (meV)  & $\epsilon_{\mathrm{HOMO}}$ (meV) & $\epsilon_{\mathrm{LUMO}}$ (meV) &  $\mu$ (D) & $C_v$ (cal/mol K) \\
\hline
Sparse Method & 0.08 & 48 & 30 & 25 & 0.29 & 0.31 \\
\hline
Gaunt Tensor Product & 0.15 & 50 & 33 & 28 & 0.33 & 0.38 \\
\hline
Fourier Transform (HEALPix) & 0.13 & 49 & 31 & 26 & 0.31 & 0.35 \\
\hline
Fourier Transform (s2fft) & 0.10 & 49 & 31 & 26 & 0.31 & 0.35 \\
\hline
\end{tabular}
}
\caption{Comparison of tensor product methods in terms of MAE on QM9 dataset. Advanced and GPU-parallelized methods are expected to yield better accuracy. The Gaunt tensor product has an issue with parity, as is explained in \cite{xie2024price}. The Sparse method has $\mathcal{O}(L^{3})$ scaling. All other methods achieve $\mathcal{O}(L^{3})$ scaling.  }
\end{table*}

\section{Model Ablations}\label{Appendix: Model Ablations}

We consider a set of ablations on our model. We hope these ablations help the reader navigate the design space of this class of models.

\subsection{Invariant embeddings}

Tasks of interest usually consisted of both invariant features and vector features. We found that using an encoder to transform all features into invariant representations improved model performance. We used Laplacian graph eigenvectors as positional features to provide a global structural signal. These eigenvectors were computed from the normalized graph Laplacian and concatenated with the input features before encoding. When additional information, such as charge in the Nbody experiment, was given, these properties were encoded using a learned one hot encoding. Learned output types are encoded through standard neural network, i.e. the encoder is a non-equivariant neural network.

\subsection{Encoding Ablation}\label{Appendix: Encoding Ablation}

We consider two ablations on the structure equivariant MLP between layers. In the main text of the paper, we used layers constructed from e3nn \cite{geiger2022e3nneuclideanneuralnetworks}, with custom tensor product. We checked that each layer was equivariant using escnn \cite{cesa2021ENsteerable}. The original $SE(3)$-Hyena paper used layers constructed from Clifford-MLP\cite{ruhe2023cliffordgroupequivariantneural}. We ablate by testing the use of Clifford-MLP \cite{ruhe2023cliffordgroupequivariantneural}. We found no substantial difference between Clifford-MLP layers and e3nn layers, indicating that the attention mechanism is the key for downstream model performance.


\begin{table}[h]
\centering
\begin{tabular}{|l|c|c|c|c|c|c|c|c|c|c|c|c|}
\hline
& \multicolumn{3}{c|}{\textbf{N=5}} & \multicolumn{3}{c|}{\textbf{N=10}} & \multicolumn{3}{c|}{\textbf{N=20}} & \multicolumn{3}{c|}{\textbf{N=40}} \\
\cline{2-13}
\textbf{Model} & $\text{mse}_x$ & $\text{mse}_v$ & $\delta \text{EQ}$ & $\text{mse}_x$ & $\text{mse}_v$ & $\delta \text{EQ}$ & $\text{mse}_x$ & $\text{mse}_v$ & $\delta \text{EQ}$ & $\text{mse}_x$ & $\text{mse}_v$ & $\delta \text{EQ}$ \\
\hline
e3nn layers & $0.0041$ & $0.0065$ & $0.0000096$ & $0.012$ & $0.019$ & $0.0000091$ & $0.026$ & $0.031$ & $0.000061$ & $0.031$ & $0.049$ & $0.000049$ \\
\hline
Clifford MLP layers & $0.0043$ & $0.0068$ & $0.0019$ & $0.011$ & $0.023$ & $0.000045$ & $0.026$ & $0.033$ & $0.000081$ & $0.035$ & $0.045$ & $0.00042$ \\
\hline
\end{tabular}
\caption{Comparison of e3nn layers vs Clifford MLP \cite{ruhe2023cliffordgroupequivariantneural}. Performance comparison across different $N$ values.}
\label{tab:performance_table;Encoders}
\end{table}

\subsection{Multiple Head Clebsch-Gordan Attention}\label{Appendix: Feature Cross Attention}

In practice, we found that using multiple heads in the tensor product lead to better performance. Specifically, query and key features $q_{i}$ and $k_{i}$ which are inputs to Clebsch-Gordan convolution are set to have $h$ independent heads. This heads do not interact with each other during the Clebsch-Gordan convolution, but are allowed to interact during linear layers and non-linearities. Thus, there are three architecture parameters to consider in a CG convolution, the number of heads $h$, the maximum harmonic $L$ and the channel dimension $m$. Query and key were always constrained to have the same number of heads, maximum harmonic and channel dimension. The total output dimension is has memory scaling $O( h \cdot L^{2} \cdot m )$. We found that performance was sensitive to the choice of $(h,L,m)$, setting the head dimension $h$ or the channel dimension $m$ too small led to a drastic decrease in performance. We found that increasing the harmonic number $L$ always improved performance, at the cost of increased memory usage.

\begin{table*}[ht]
\centering
\caption{Mean absolute error (MAE) on QM9 dataset for different maximum spherical harmonic degrees \(L\). Lower is better.}
\label{tab:qm9_harmonics_results}
\setlength{\tabcolsep}{3pt}
\fontsize{8pt}{3pt}\selectfont

\resizebox{\textwidth}{!}{
\begin{tabular}{|l|c|c|c|c|c|c|}
\hline
\textbf{Max Harmonic Degree \(L\)} &  $\alpha$ (Bohr$^3$) & $\Delta \epsilon$ (meV)  & $\epsilon_{\mathrm{HOMO}}$ (meV) & $\epsilon_{\mathrm{LUMO}}$ (meV) &  $\mu$ (D) & $C_v$ (cal/mol K) \\
\hline
\(L = 3\) & 0.53 & 68 & 71 & 59 & 0.93 & 0.89 \\
\hline
\(L = 4\) & 0.25 & 59 & 42 & 35 & 0.69 & 0.58 \\
\hline
\(L = 5\) & 0.19 & 51 & 33 & 29 & 0.50 & 0.39 \\
\hline
\(L = 6\) & 0.10 & 49 & 31 & 26 & 0.31 & 0.35 \\
\hline
\end{tabular}
}
\caption{Model Ablation: Comparison of performance on QM9 as a function of the maximum harmonic $L$ used. Performance monotonically increases with the maximum harmonic $L$. }
\end{table*}

\subsection{Gating Ablation}\label{Append: Gating Abalation}

The gating mechanism is the key component in transformers. In our proposed method, we perform another tensor product to combine queries, keys and values. However, it is not obvious if this is needed. As an alternative, it maybe possible to simply concatenate $u_{i}^{\ell}$ with values $v_{i}^{\ell}$. We found that this leads to slight decrease in performance (although decreases the forward pass time by a sizable margin). 

\begin{table*}[ht]
\centering
\caption{Mean absolute error (MAE) on QM9 dataset for different gating types. Lower is better }
\label{tab:qm9_harmonics_results}
\setlength{\tabcolsep}{3pt}
\fontsize{8pt}{3pt}\selectfont

\resizebox{\textwidth}{!}{
\begin{tabular}{|l|c|c|c|c|c|c|}
\hline
\textbf{Gating Type} &  $\alpha$ (Bohr$^3$) & $\Delta \epsilon$ (meV)  & $\epsilon_{\mathrm{HOMO}}$ (meV) & $\epsilon_{\mathrm{LUMO}}$ (meV) &  $\mu$ (D) & $C_v$ (cal/mol K) \\
\hline
Concatenation gating & 0.25 & 61 & 46 & 31 & 0.63 & 0.51 \\
\hline
CG gating & 0.10 & 49 & 31 & 26 & 0.31 & 0.35 \\
\hline
\end{tabular}
}
\caption{Model Ablation: Comparison of performance on QM9 as a function of the maximum harmonic $L$ used. Models were trained with $L=6$ and four heads. }
\end{table*}

\end{document}